# Opinion Polarity Identification through Adjectives

Samaneh Moghaddam  
School of Computing Science  
Simon Fraser University  
Burnaby, BC, Canada  
**sam39@cs.sfu.ca**

Fred Popowich  
School of Computing Science  
Simon Fraser University  
Burnaby, BC, Canada  
**popowich@sfu.ca**

## ABSTRACT

"What other people think" has always been an important piece of information during various decision-making processes. Today people frequently make their opinions available via the Internet, and as a result, the Web has become an excellent source for gathering consumer opinions. There are now numerous Web resources containing such opinions, e.g., product reviews forums, discussion groups, and Blogs. But, due to the large amount of information and the wide range of sources, it is essentially impossible for a customer to read all of the reviews and make an informed decision on whether to purchase the product. It is also difficult for the manufacturer or seller of a product to accurately monitor customer opinions. For this reason, mining customer reviews, or opinion mining, has become an important issue for research in Web information extraction. One of the important topics in this research area is the identification of opinion polarity. The opinion polarity of a review is usually expressed with values 'positive', 'negative' or 'neutral'. We propose a technique for identifying polarity of reviews by identifying the polarity of the adjectives that appear in them. Our evaluation shows the technique can provide accuracy in the area of 73%, which is well above the 58%-64% provided by naïve Bayesian classifiers.

## 1. INTRODUCTION

The Web has significantly changed the way that consumers express their opinions. Today product reviews exist in a variety of forms on the Web; however, reading a representative selection of product reviews to make a good decision is frequently a time-consuming process. It is also difficult for the producer or seller of a product to make use of the reviews. Recently, there has been interest in the development and use of mining and summarization tools specifically for identifying opinion polarity [1][2].

The polarity of a word or opinion, also called semantic orientation, indicates the direction the word or opinion deviates from the norm for its semantic group. In other words, the polarity of the words or phrases directly shows the opinion of the text writer. Polarity usually ranges on over an ordinal (i.e., discrete) scale. This scale may be in the form either of an ordered set of numerical values (e.g., one to five "stars"), or of an ordered set of non-numerical labels (e.g., positive, negative, neutral); the only difference between these two cases is that in the former case the distances between consecutive scores are known while in the latter the distances are not known.

In this paper we propose an algorithm for identifying the polarity of customer reviews. This algorithm has three phases: extract adjectives and their frequencies from the given review, predict the polarity of each adjective using the learned classifier, and classify the review based on the polarity of the adjectives.

The remainder of the paper is organized as follows. The next section is concerned with the proposed technique and its phases, including 'similarity computation', 'polarity classification' and 'opinion polarity identification'. In Section 3 we will explain the experimental results and the evaluation of our method. Sections 4 reviews related work and the last section provides conclusions and discusses future work.

## 2. PROPOSED TECHNIQUE

Given a review, our algorithm first collects all of the adjectives from the review and then computes the frequency of each of them. In the next step, it predicts the polarity of each adjective using a learned classifier. Then by aggregating the polarity of the opinion's adjectives (based on their frequencies), the polarity of the opinion is identified.

For determining the polarity of an adjective, we use a naïve Bayesian classifier which was learned by a set of positive, negative, and neutral adjectives. The input of this classifier is a tuple of values showing the similarity between the given adjective and three fixed adjectives 'excellent', 'mediocre' and 'poor' as representative of positive, neutral and negative adjectives. These similarity values are computed using two proposed similarity functions which are based on WordNet [3]. The first similarity function is based on the adjectives' synonyms and the second one is based on the stem of the adjectives and a predefined similarity function for nouns and verbs. In the following we explain each of these phases in detail.

## 2.1 SIMILARITY COMPUTATION

In this phase we define a similarity function for computing similarity between two adjectives making use of the WordNet functions provided in NLTK [4]. While some similarity measures have been provided for working over collections of nouns and verbs, these similarity measures do not work for adjectives and adverbs, since these functions compute similarity between two words by using hyponym/hypernym (or is-a relations). Due to the limitation of is-a hierarchies which are only available in WordNet for nouns and verbs, the defined similarity functions only work with noun-noun and verb-verb parts of speech.

The first similarity function we define for adjectives is called *SynSim* which is based on three different features of a WordNet adjective: 'synonym' words, 'similar to' words and 'see also' words. All of these features contain words (mainly adjectives) having the same or nearly the same meaning.

We take advantage of all three features and consider the words they contain as synonym words but with different weights. For example, in WordNet 'good' and 'excellent' are not listed as synonyms. They are listed as 'similar' words. So, if we just use synonym words we will lose a great deal of possibly appropriate words.

In addition, since 'synonymy' shows a stronger relation than 'similarity', and since 'similar words' are stronger than 'see also' words, we assign different weights to them. We apply the weights of 1, 0.9 and 0.8 for 'synonym' words, 'similar to' words and 'see also' words respectively. Ultimately, the best weight values can be found experimentally.

We can now define *SynSim* as a recursive method in which the similarity between $adj_1$ and $adj_2$ is computed using the similarity value between a synonym of $adj_1$ and $adj_2$. We propose a maximum recursion level of three for our initial experiments to control the running time. Table 1 shows some pairs of adjectives and their similarity values computed by the *SynSim* function.

**Table 1: Similarity values using SynSim**

| Word 1 | Word 2 | SynSim |
|---|---|---|
| amazing | awesome | 1.00 |
| pretty | attractive | 0.36 |
| ugly | beautiful | 0 |
| assistant | helpful | 0.22 |
| mediocre | average | 1.00 |
| bad | excellent | 0 |
| attractive | appealing | 0 |
| readable | fine | 0.20 |
| blurry | foggy | 1.00 |

The other similarity function, called *StemSim*, is based on one of the predefined similarity functions in NLTK for nouns and verbs (`path_similarity`). The *StemSim* first determines the stem of the given adjective which is usually a noun or verb. Of course the stem of some of the adjectives are themselves adjectives, like 'poor'. In this case, the method searches for a noun or verb in the synonym list and use it instead of the stem. In the next step, *StemSim* uses the predefined path-similarity function to compute the similarity of two corresponding stems. Table 2 shows the *StemSim* similarity values for some adjectives.

**Table 2: Similarity values using StemSim**

| Word 1 | Word 2 | StemSim |
|---|---|---|
| amazing | awesome | 1.00 |
| pretty | attractive | 0.22 |
| ugly | beautiful | 0 |
| assistant | helpful | 1.00 |
| mediocre | average | 1.00 |
| bad | excellent | 0 |
| attractive | appealing | 0.10 |
| readable | fine | 0.07 |
| blurry | foggy | 1.00 |

Comparing the results of the similarity functions shows that computing similarities based on synonym words is more accurate than computing similarities based on the stems of words. However, we use both functions in the next phases and compare the final results together.

In this subsection, we explained how to compute the similarity of two adjectives using synonyms or corresponding stems. In the next subsection we will explain how to use similarity values to determine the polarity of adjectives.

## 2.2 POLARITY CLASSIFICATION

As we explained before, polarity of a word refers to its strength in a classification, typically in a 'positive' vs. 'negative' sense. In this work we consider three polarity levels for adjectives: positive, negative and neutral. An adjective can imply positive meaning, like 'excellent', negative meaning, like 'poor' and neutral meaning, like 'mediocre'.

For determining the polarity of adjectives we use a naïve Bayesian classifier to classify adjectives. We use a set of tagged adjectives for training and testing the classifier. In this set there are 30 adjectives, 10 of them are tagged as positive, 10 as negative and 10 as neutral adjectives (Table 3). We randomly select 15 adjectives for training the classifier and use the remaining as test set.

The input of naïve Bayesian classifier is a set of tuples containing the similarity values between the given adjective and the three fixed adjectives 'excellent', 'mediocre' and 'poor'. We use these three fixed adjectives as reference points to predict the polarity of other adjectives. These similarity values are computed using the two proposed similarity functions, *SynSim* and *StemSim*, which were introduced in Section 2.1. Table 4 shows the input for learning a classifier.

**Table3: Tagged adjectives used for training and testing the classifier**

| Positive | Neutral | Negative |
|---|---|---|
| good | mediocre | bad |
| nice | average | awful |
| awesome | enough | defective |
| excellent | fair | faulty |
| great | okay | poor |
| perfect | ordinary | unsatisfactory |
| precious | fine | imperfect |
| satisfactory | suitable | weak |
| exceptional | reasonable | bitter |
| outstanding | neutral | terrible |

**Table 4: Sample input for learning Naïve Bayesian classifier. The input is a set of tuples containing similarity values to (excellent, mediocre, poor) computed by SynSim and StemSim.**

| Adjective | SynSim | StemSim |
|---|---|---|
| nice | (1,0,0) | (0,0,0) |
| weak | (0,0,1) | (0,1,0) |
| bitter | (0,1,1) | (0,0,1) |
| neutral | (0,1,0) | (0,1,0) |
| bad | (0,1,1) | (0,0,1) |
| perfect | (0,1,0) | (0,1,0) |
| great | (0,1,1) | (0,1,0) |
| … | | |

As shown in Table 4, the method rounds the similarity values to be able to learn the classifier with very few cases. Since the training set only has 15 tagged adjectives, it is very hard to learn the classifier by using the actual values. Details on the rounding process will not be introduced here due to space constraints.

Given an adjective from the training set, for example 'nice', the method computes three similarity values, excellent_sim = SynSim('nice','excellent') = 0.16, mediocre_sim = SynSim ('nice', 'mediocre') = 0, and poor_sim = SynSim ('nice', 'poor') = 0. Then it rounds them and sends them as a tuple (excellent_sim, mediocre_sim, poor_sim) to learn the classifier.

The accuracy of the learned classifiers (based on different similarity functions) and the most informative features for each is provided in Tables 5 and 6. Note that, the accuracy and important features of the classifiers may vary in different runs, since the training set is randomly selected.

**Table 5: Accuracy and important features of SynClassifier (based on SynSim function)**

| Accuracy | 0.8 | |
|---|---|---|
| Important features | poor = 1 | neg : nut = 4.3 : 1.0 |
| | mediocre = 1 | neg : pos = 1.8 : 1.0 |
| | mediocre = 0 | pos : neg = 1.7 : 1.0 |
| | excellent=0 | nut : pos = 1.6 : 1.0 |
| | poor = 0 | pos : nut = 1.1 : 1.0 |

**Table 6: Accuracy and important features of StemClassifier (based on StemSim function)**

| Accuracy | 0.46 | |
|---|---|---|
| Important features | mediocre = 0 | neg : nut = 2.8 : 1.0 |
| | mediocre = 1 | nut : neg = 2.5 : 1.0 |
| | poor = 0 | nut : neg = 2.0 : 1.0 |
| | poor = 1 | neg : nut = 1.7 : 1.0 |
| | excellent = 0 | pos : neg = 1.0 : 1.0 |

As shown in Table 6 the accuracy of *StemClassifier* is almost the same as the accuracy of random classification (about 0.33 for three classes). So using the stems of adjectives is not a good method for classifying them. On the other hand, the accuracy of *SynClassifier* which is learned using *SynSim* function is high. The *SynClassifier* can correctly classify a new adjective as positive, negative, or neutral with the probability of 80% which is a high accuracy value in this area.

In addition, as we explained before, the size of the training set we used here is small (just 15 cases). So, it is mostly probable that if we increase the size of the training set, the accuracy of the classifier will be increased. In the following phases we will use both classifiers.

After learning the classifier, the method is able to predict the polarity of new adjectives. Table 7 shows some sample adjectives which are classified by the learned classifiers. As we expected, the predicted polarities by the *SynClassifier* are more accurate than those by the *StemClassifier*.

**Table 7: Predicted polarity for new adjectives.**

| Adjective | SynClf Polarity | StemClf Polarity |
|---|---|---|
| beautiful | positive | positive |
| ugly | negative | negative |
| incorrect | positive | neutral |
| ordinary | neutral | neutral |
| fabulous | positive | neutral |
| worthless | negative | negative |
| conservative | neutral | neutral |
| adequate | positive | neutral |
| nice | positive | positive |
| … | | |

So far the method can identify the polarity of a single adjective. In the next step it tries to predict the polarity of an opinion using the polarity of the adjectives appearing in the opinion.

## 2.3 POLARITY IDENTIFICATION

The last phase of our technique involves determining opinion polarity. The polarity is determined by aggregating the polarity of the extracted adjectives based on their frequencies. In other words, for each review our method assigns the scores +1, 0, and -1 to the positive, neutral, and negative adjectives respectively. Each adjective is also assigned a weight which is equal to its frequency in that review. The polarity of the review is then determined by computing the weighted average of the adjective scores. We consider two methods, the first, *SynPI*, is based on *SynClassifier*, and the second, *StemPI*, is based on *StemClassifier*.

Table 8 shows the accuracy, false-positive and false-negative percentages of the two polarity identification methods *SynPI* and *StemPI*. Note again that since the test set is selected randomly, the results may vary in different runs.

**Table 8: Accuracy, false-positive and false-negative of predicted polarity for 100 reviews using two polarity identifier (SynPI, StemPI).**

| Method | Accuracy | False-positive | False-negative |
|---|---|---|---|
| SynPI | 73% | 2% | 25% |
| StemPI | 59% | 37% | 4% |

As shown in Table 8, the accuracy of opinion polarity classification methods, *SynPI* and *StemPI*, are 73% and 59% respectively. It shows that, as we expected, *StemPI* method does not perform well for opinion classification. On the other hand, the *SynPI* identifies the polarity of reviews with high accuracy which shows the power of the learned classifier (*SynClassifier*).

In the next section we will evaluate our proposed technique by comparing it with a baseline classification method.

## 3. EVALUATION

In this section, we evaluate the accuracy of our proposed opinion polarity classifier, *SynPI*, by comparing it with a baseline classification method (presented in Chapter 6 of NLTK [4]). The baseline classification method is based on a naïve Bayesian classifier learned using a set of tagged reviews. In order to evaluate our system, we use a set of reviews available via `movie_reviews` in `nltk.corpus` [4]. These reviews have been categorized into 'positive' and 'negative' classes. Our evaluation technique involves the random selection of a set of 100 reviews, which are then provided to our system for polarity prediction.

This baseline method first finds the most frequent words in the training set, and then uses them to learn the naïve Bayesian classifier. Each feature of this classifier indicates whether the review contains a special word [4].

Table 9 shows accuracy and the first five important features of BaselineClassifier for different sizes of training sets (25, 50 and 100). The size of the test set which is randomly selected in each run is 100.

As shown in Table 9, the accuracy of the baseline method is increased by increasing the size of the training set. By using 25 reviews in a training set the method does not perform better than random classification, and by using a training set with 100 reviews, we get the accuracy of 64%. Of course, as mentioned in the NLTK book, if we use the whole dataset (containing 2000 reviews) as a training set and just use 100 reviews for test, the accuracy will be about 80%. However, producing sufficient quantities of labeled data can be very expensive in manual effort.

**Table 9: Accuracy and important features of BaselineClassifier (based on StemSim function)**

| Size | Acc | Important features | |
|---|---|---|---|
| 25 | 58% | has(help) = True | pos : neg =4.0 : 1.0 |
| | | has(going) = True | pos : neg =4.0 : 1.0 |
| | | has(series) = True | pos : neg =4.0 : 1.0 |
| | | has(acting) = True | neg : pos =3.9 : 1.0 |
| | | has(friend) = True | neg : pos =3.9 : 1.0 |
| 50 | 60% | has(performances)=True | pos : neg =5.4 : 1.0 |
| | | has(discovers) = True | pos : neg =5.4 : 1.0 |
| | | has(girl) = True | neg : pos =5.0 : 1.0 |
| | | has(search) = True | pos : neg =4.3 : 1.0 |
| | | has(score) = True | pos : neg =4.3 : 1.0 |
| 100 | 64% | has(lame) = True | neg : pos =7.4 : 1.0 |
| | | has(definitely) = True | pos : neg =5.8 : 1.0 |
| | | has(wasted) = True | neg : pos =5.7 : 1.0 |
| | | has(guess) = True | neg : pos =5.7 : 1.0 |
| | | has(agent) = True | neg : pos =5.7 : 1.0 |

We can see that, the accuracy of the baseline method when it is trained with 100 reviews is lower than our proposed method which required just 15 adjectives in its training. This is the first advantage of our proposed method which is that it is able to provide high accuracy while using a small training set, as summarized in Table 10.

**Table 10: Accuracy evaluation.**

| Method | Accuracy |
|---|---|
| SynPI | 73% |
| BasPI-25 | 58% |
| BasePI-50 | 60% |
| BasePI-100 | 64% |

Another advantage of our method is its independency from the context. The baseline method learns based on the frequency of words in the given reviews. So, it totally depends on the context of the training dataset. As shown in Table 9, the important features of the classifier are totally dependent on the review context. So, for each context we need a specific classifier. On the other hand, our proposed method is independent from the context and it depends only on the adjectives that appear in the text. When our method learns the opinion classifier, it can be applied to other contexts.

## 4. RELATED WORK

Polarity and subjectivity of words has been studied previously in two main classes of research. One class consists of work exclusively aimed at the determination of effective metrics for representing the polarity and subjective content of words [5][6]. The other class consists of work that performs word polarity as part of a larger investigation into opinion classification and related domains [1]. Our work lies in the second class, so we focus on the opinion classification in our related work.

Turney in [1] introduces an unsupervised learning algorithm for rating a review as positive or negative (thumbs up or down). The algorithm extracts phrases containing adjectives or adverbs and also their Part Of Speech (POS) tags conform to any of the predefined patterns. This paper also proposes a method 'PMI-IR' for computing semantic orientation of the phrases. PMI-IR uses Pointwise Mutual Information (PMI) and Information Retrieval (IR) to measure the similarity of pairs of words or phrases. Finally, by averaging the polarity of the extracted phrases, the method classifies the review as positive or negative. Our proposed method is somewhat similar to Turney's method, except that we use three polarities, and require minimal training data.

Dave et al. in [7] describe a tool for sifting through and combining product reviews. They used structured reviews for training and testing. The method first selects features using some substitution methods, and then uses the proposed score function to determine the polarity of them. The proposed scoring method used some machine-learning techniques using the Rainbow text-classification package to identify the polarity of each feature. Finally it uses a trained classifier to classify the reviews as positive or negative. Our approach obtains similar performance, but again requires less in the way of training information.

Kamps et al. in [8] investigate Osgood's measures based on the WordNet knowledge database. Using the adjectives 'good' and 'bad' to calculate semantic polarity of word is the main evaluative dimension of Osgood. The minimal distance $d(w_i,w_j)$ between words and 'good' or 'bad' shows the similarity of their semantic orientation. So, like our work, they were influenced by WordNet, but we have the further distinction of neutral polarity.

In the same way, Yu et al. in [9] use the HowNet knowledge database to investigate Osgood's measures to see whether this approach can be used for Chinese. In this paper, sentiment features of text are divided into characteristic words and phrases, which are extracted from the training data. Their method combines HowNet with a sentiment classifier to compute semantic similarity of characteristic words with tagged words in HowNet. It adopts the positive and negative terms as features of a sentiment classifier.

Pang et al. [10] use three machine-learning methods (Naïve Bayes, Maximum Entropy classification, and Support Vector Machines) for classifying reviews as positive or negative and show that the standard machine learning techniques outperform human-produced baselines. The results also show that the SVM technique works better in comparison with the others, but requires a substantial amount of training information.

## 5. CONCLUSION

This paper introduces a technique for classifying a review as positive, negative, or neutral. The algorithm has three phases: (1) extracting adjectives and their frequencies from the given review, (2) predicting the polarity of each adjective using the learned classifier, and (3) classifying the review based on the weighted average polarity of the adjectives. The core of the technique is the second phase, which uses WordNet to compute similarity values between two adjectives, and then uses the similarity values to learn the classifier for predicting polarity of each adjective.

In experiments with 100 reviews from the `movie_reviews` corpus in NLTK, our algorithm attains an accuracy of 73% while the baseline method in the best case can attain an accuracy of 64%. The key advantage of our method is that it can attain high accuracy using a small training set.

In addition, our proposed opinion polarity classifier is independent from the context and can be applied to different review types. On the other hand, the baseline classifier totally depends on the context which limits its usage. In the end, we can say that high accuracy along with simplicity of our method may encourage further work with opinion polarity.

## REFERENCES

[1] Turney P. D. "Thumbs Up or Thumbs Down? Semantic Orientation Applied to Unsupervised Classification of Reviews", ACL 2002.

[2] Salvetti F., Lewis S. and Reichenbach C., "Automatic Opinion Polarity Classification of Movie Reviews", Colorado Research in Linguistics. Volume 17, Issue 1, June 2004.

[3] Fellbaum, C., WordNet: An Electronic Lexical Database. Bradford Books, 1998.

[4] Bird, S. and and Loper, E.. Proceedings of the ACL demonstration session. pp 214-217, Barcelona, Association for Computational Linguistics, July 2004.

[5] Baroni M. and Vegnaduzzo S., "Identifying Subjective Adjectives through Web-based Mutual Information", In Ernst Buchberger (ed.), Proceedings of KONVENS, Vienna: GAI. 17-24, 2004.

[6] Vegnaduzzo S., "Acquisition of subjective adjectives with limited resources", In AAAI Spring Symposium Technical Report: Exploring Affect and Attitude in Text, 2004.

[7] Dave K., Lawrence S. And Pennock D. M., "Mining the peanut gallery opinion extraction and semantic classification of product reviews", WWW2003.

[8] Kamps J., Marx M., Mokken R., Rijke m., "Using WordNet to Measure Semantic Orientations of Adjectives", Proceedings of the fourth international conference on Language Resources and Evaluation, pp 1115-1118, Lisbon, Portugal, 2004.

[9] Yu L., Ma J., Tsuchiya S. and Ren F., "Opinion Mining: A Study on Semantic Orientation Analysis for Online Document", Proceedings of the 7th World Congress on Intelligent Control and Automation, Chongqing, China, June 25 - 27, 2008.

[10] Pang B., Lee L. and Vaithyanathan S., "Thumbs up? Sentiment Classification using Machine Learning Techniques", ACL-02 conference on Empirical methods in natural language processing - Volume 10, PP 79-86, 2002.